%% file: naaclhlt2019.tex
\documentclass[11pt,a4paper]{article}
\usepackage[hyperref]{naaclhlt2019}
\usepackage{times}
\usepackage{latexsym}

\usepackage{url}
\usepackage{multirow}
\usepackage{adjustbox}
\usepackage{graphicx}
\usepackage[scaled=.90]{helvet}
\usepackage{amsfonts}
\usepackage{float}
\restylefloat{table}

\aclfinalcopy 
\def\aclpaperid{449} 

\newcommand\blfootnote[1]{%
\begingroup 
\renewcommand\thefootnote{*}\footnotetext{#1}%
\addtocounter{footnote}{0}%
\endgroup 
}

\newcommand\bifootnote[1]{%
\begingroup 
\renewcommand\thefootnote{$\dag$}\footnotetext{#1}%
\addtocounter{footnote}{0}%
\endgroup 
}

\title{Microblog Hashtag Generation via Encoding Conversation Contexts}

\author{Yue Wang $^{1*}$
~Jing Li$^{2\dag}$ ~Irwin King$^{1}$ ~ Michael R. Lyu$^{1}$~ Shuming Shi$^{2}$\\
$^{1}$Department of Computer Science and Engineering \\
The Chinese University of Hong Kong, HKSAR, China \\
$^{2}$Tencent AI Lab, Shenzhen, China\\
$^{1}$\texttt{\{yuewang, king, lyu\}@cse.cuhk.edu.hk}\\
$^{2}$\texttt{\{ameliajli,shumingshi\}@tencent.com}\\}

\date{}

\begin{document}
\maketitle
\blfootnote{This work was mainly done when Yue Wang was an intern at Tencent AI Lab.}
\bifootnote{Jing Li is the corresponding author.}
\input{sections/abstract}

\input{sections/introduction}
\input{sections/model}
\input{sections/exp-setup}

\input{sections/exp-result}
\input{sections/related-work}
\input{sections/conclusion}
\input{sections/ack.tex}

\bibliography{naaclhlt2019}
\bibliographystyle{acl_natbib}

\end{document}

%% file: sections/abstract.tex
\begin{abstract}
Automatic hashtag annotation plays an important role in content understanding for microblog posts. 
To date, progress made in this field has been restricted to phrase selection from limited candidates, or word-level hashtag discovery using topic models.
Different from previous work considering hashtags to be inseparable, our work is the first effort to annotate hashtags with a novel \textit{sequence generation} framework via viewing the hashtag as a short sequence of words.  
Moreover, to address the data sparsity issue in processing short microblog posts, we propose to jointly model the target posts and the \textit{conversation contexts} initiated by them with bidirectional attention.
Extensive experimental results on two large-scale datasets, newly collected from English Twitter and Chinese Weibo, show that our model significantly outperforms state-of-the-art models based on classification.\footnote{Our code and data are released in \url{https://github.com/yuewang-cuhk/HashtagGeneration}}
Further studies demonstrate our ability to effectively generate rare and even unseen hashtags, which is however not possible for most existing methods.
\end{abstract}

%% file: sections/introduction.tex
\section{Introduction}

Microblogs have become an essential outlet for individuals to voice opinions and exchange information. 
Millions of user-generated messages are produced every day, far outpacing the human being's reading and understanding capacity. 
As a result, the current decade has witnessed the increasing demand for 
effectively discovering gist information from large microblog texts.
To identify the key content of a microblog post, hashtags, user-generated labels prefixed with a ``\textit{\#}'' (such as ``\textit{\#NAACL}'' and ``\textit{\#DeepLearning}''), have been widely used to reflect keyphrases~\cite{DBLP:conf/emnlp/ZhangWGH16,DBLP:conf/naacl/ZhangLSZ18} or topics~\cite{DBLP:conf/www/YanGLC13,DBLP:conf/www/HongAGST12,DBLP:conf/acl/LiLGHW16}. 
Hashtags can further benefit downstream applications, such as microblog search~\cite{DBLP:conf/sigir/Efron10,DBLP:conf/www/BansalJV15}, summarization~\cite{DBLP:journals/taslp/ZhangLGY13,DBLP:conf/wsdm/ChangWML13}, 
sentiment analysis~\cite{DBLP:conf/coling/DavidovTR10,DBLP:conf/cikm/WangWLZZ11}, and so forth. 
Despite the widespread use of hashtags, there are a large number of microblog messages without any user-provided hashtags. 
For example, less than $15$\% tweets contain at least one hashtag~\cite{DBLP:conf/cikm/WangWLZZ11,DBLP:conf/ht/KhabiriCK12}.
Consequently, for a multitude of posts without human-annotated hashtags, there exists a pressing need for automating the hashtag annotation process for them.

\input{tabs/intro_example.tex}

Most previous work in this field focuses on extracting phrases from target posts~\cite{DBLP:conf/emnlp/ZhangWGH16,DBLP:conf/naacl/ZhangLSZ18} or selecting candidates from a pre-defined list~\cite{DBLP:conf/ijcai/GongZ16,DBLP:conf/coling/HuangZGH16,DBLP:conf/ijcai/ZhangWHHG17}. 
However, hashtags usually appear in neither the target posts nor the given candidate list. 
The reasons are two folds.
For one thing, microblogs allow large freedom for users to write whatever hashtags they like.
For another, due to the wide range and rapid change of social media topics, a vast variety of hashtags can be daily created, making it impossible to be covered by a fixed candidate list.
Prior research from another line employs topic models to generate topic words as hashtags ~\cite{DBLP:conf/emnlp/GongZH15,DBLP:conf/emnlp/ZhangWGH16}. These methods, ascribed to the limitation of most topic models, are nevertheless incapable of producing phrase-level hashtags.

In this paper, we approach hashtag annotation from a novel \emph{sequence generation} framework. 
In doing so, we enable phrase-level hashtags beyond the target posts or the given candidates to be created. 
Here, hashtags are first considered as a sequence of tokens (e.g., ``\textit{\#DeepLearning}'' as ``\textit{deep learning}''). 
Then, built upon the success of sequence to sequence (seq2seq) model on language generation~\cite{ DBLP:conf/nips/SutskeverVL14}, we present a neural seq2seq model to generate hashtags in a \textit{word-by-word} manner.
To the best of our knowledge, \emph{we are the first to deal with hashtag annotation in sequence generation architecture}. 

In processing microblog posts, one major challenge we might face is the limited features to be encoded. It is mostly caused by the data sparsity exhibited in short and informal microblog posts.\footnote{For instance, the eligible length of a post on Twitter or Weibo is up to $140$ characters.} 
To illustrate such challenge, Table \ref{tab:intro-example} displays a sample Twitter post tagged with ``\textit{\#AusOpen}'', referring to Australian Open tennis tournament.
Only given the short post, it is difficult to understand why it is tagged with ``\textit{\#AusOpen}'', not to mention that neither ``\textit{aus}'' nor ``\textit{open}'' appear in the target post. In such a situation, how shall we generate hashtags for a post with limited words?

To address the data sparsity challenge, we exploit conversations initiated by the target posts to enrich their contexts. 
Our approach is benefited from the nature that most messages in a conversation tend to focus on relevant topics. 
Content in conversations might hence provide contexts facilitating the understanding of the original post~\cite{DBLP:conf/wsdm/ChangWML13,DBLP:conf/emnlp/LiGWPW15}. 
The effects of conversation contexts, useful on topic modeling~\cite{DBLP:conf/acl/LiLGHW16,DBLP:journals/cl/LiSWW2018} and keyphrase extraction~\cite{DBLP:conf/naacl/ZhangLSZ18}, have never been explored on microblog hashtag generation.
To show why conversation contexts are useful, we display in Table \ref{tab:intro-example} a conversation snippet formed by some replies of the sample target post. 
As can be seen, key content words in the conversation (e.g., ``\textit{Nadal}'', ``\textit{Tomic}'', and ``\textit{tennis}'') are useful to reflect the relevance of the target post to the hashtag ``\textit{\#AusOpen}'', because Nadal and Tomic are both professional tennis players.
Concretely, our model employs a dual encoder (i.e., two encoders), one for the target post and the other for the conversation context, to capture the representations from the \textit{two sources}.
Furthermore, to capture their joint effects, we employ the bidirectional attention (\textbf{bi-attention})~\cite{DBLP:journals/corr/SeoKFH16} to explore the interactions between two encoders' outputs.
Afterward, an attentive decoder is applied to generate the word sequence of the hashtag. 

In experiments, 
we construct two large-scale datasets, one from English platform Twitter and the other from Chinese Weibo.
Experimental results based on both information retrieval and text summarization metrics show that our model generates hashtags closer to human-annotated ones than all the comparison models. 
For example, our model achieves $45.03$\% ROUGE-1 F1 on Weibo, compared to $25.34$\% given by the state-of-the-art classification-based method. 
Further comparisons with classification-based models show that our model, in a sequence generation framework, can better produce rare and even new hashtags. 

To summarize, our contributions are three-fold:

$\bullet$~We are the first to approach microblog hashtag annotation with \emph{sequence generation} architecture.

$\bullet$~To alleviate data sparsity, we enrich context for short target posts with their \emph{conversations} and employ a bi-attention mechanism for capturing their interactions.

$\bullet$~Our proposed model outperforms state-of-the-art models by large margins on two large-scale datasets, constructed as part of this work.

%% file: tabs/intro_example.tex
\begin{table}[t]
\centering
\scalebox{0.95}{
\begin{tabular}{|p{7.6cm}|}
\hline
\underline{\textbf{Target post for hashtag generation}}\\
This \textcolor{blue}{\textit{Azarenka}} woman needs a talking to from the umpire her weird noises are totes inappropes professionally. \textit{\textbf{\#AusOpen}}\\
\hline
\hline
\underline{\textbf{Replying messages forming a conversation}}\\
\textbf{[T1]} How annoying is she. I just worked out what she sounds like one of those turbo charged cars when they change gear or speed.\\

\textbf{[T2]} On the topic of noises, I was at the \textcolor{blue}{\textit{NadalTomic}} game last night and I loved how quiet \textcolor{blue}{\textit{Tomic}} was compared to \textcolor{blue}{\textit{Nadal}}.\\ 

\textbf{[T3]} He seems to have a shitload of talent and the postmatch press conf. He showed a lot of maturity and he seems nice. \\

\textbf{[T4]} \textcolor{blue}{\textit{Tomic}} has a fantastic \textcolor{blue}{\textit{tennis}} brain... \\
\hline
\end{tabular}
}
\caption{A post and its conversation snippet about ``Australian Open'' on Twitter. 
``\textit{\#AusOpen}'' is the human-annotated hashtag for the target post. 
\textit{\textcolor{blue}{Words indicative of the hashtag}} are in blue and italic type.}\label{tab:intro-example}
\end{table}

%% file: sections/model.tex
\section{Neural Hashtag Generation Model}

In this section, we describe our framework shown in Figure~\ref{fig:bi_attn}. 
There are two major modules: a dual encoder to encode both target posts and their conversations with a bi-attention to explore their interactions, and a decoder to generate hashtags. 

\paragraph{Input and Output.} 
Formally, given a target post $\mathbf{x}^p$ formulated as word sequence $\langle x^p_1, x^p_2, ..., x^p_{|\mathbf{x}^p|}\rangle$ and its conversation context $\mathbf{x}^c$ formulated as word sequence $\langle x^c_1, x^c_2, ..., x^c_{|\mathbf{x}^c|}\rangle$, where $|\mathbf{x}^p|$ and $|\mathbf{x}^c|$ denote the number of words in the input target post and its conversation, respectively, our goal is to output a hashtag $\mathbf{y}$ represented by a word sequence $\langle y_{1}, y_{2}, ..., y_{|\mathbf{y}|}\rangle$. 
For training instances tagged with multiple gold-standard hashtags, we copy the instances multiple times, each with one gold-standard hashtag following \citet{DBLP:conf/acl/MengZHHBC17}. 
All the input target posts, their conversations, and the hashtags share the same vocabulary $V$.

\input{figure_input/framework.tex}

\paragraph{Dual Encoder.}
To capture representations from both target posts and conversation contexts, we design a dual encoder, composed of a post encoder and a conversation encoder, each taking the $\mathbf{x}^p$ and $\mathbf{x}^c$ as input, respectively. 

For the post encoder, we use a bidirectional gated recurrent unit (Bi-GRU)~\cite{DBLP:conf/emnlp/ChoMGBBSB14} to encode the target post $\mathbf{x}^p$, where its embeddings $e(\mathbf{x}^p)$ are mapped into hidden states $\mathbf{h}^p=\langle \mathbf{h}^p_1,\mathbf{h}^p_2,...,\mathbf{h}^p_{|\mathbf{x}^p|}\rangle$. 
Specifically, $\mathbf{h}^p_i=[\overrightarrow{\mathbf{h}^p_i};\overleftarrow{\mathbf{h}^p_i}]$ is the concatenation of forward hidden state $\overrightarrow{\mathbf{h}^p_i}$ and backward hidden state $\overleftarrow{\mathbf{h}^p_i}$ for the $i$-th token:
\begin{equation} 
\overrightarrow{\mathbf{h}^p_i}=GRU(e(\mathbf{x}^p_i),\overrightarrow{\mathbf{h}^p_{i-1}}),
\end{equation}
\begin{equation} 
\overleftarrow{\mathbf{h}^p_i}=GRU(e(\mathbf{x}^p_i),\overleftarrow{\mathbf{h}^p_{i+1}}).
\end{equation}
Likewise, the conversation encoder converts conversations into hidden states $\mathbf{h}^c$ via another Bi-GRU. 
The dimensions of both $\mathbf{h}^p$ and $\mathbf{h}^c$ are $d$. 

\paragraph{Bi-attention.}
To further distill useful representations from our two encoders, we employ the bi-attention to explore the interactions between the target posts and their conversations.
The adoption of bi-attention is inspired by~\citet{DBLP:journals/corr/SeoKFH16}, where the bi-attention was applied to extract query-aware contexts for machine comprehension.
Our intuition is that the content concerning the key points in target posts might have their relevant words frequently appearing in their conversation contexts, and vice versa.
In general, such content can reflect what the target posts focus on and hence effectively indicate what hashtags should be generated.
For instance, in Table \ref{tab:intro-example}, names of tennis players (e.g., ``\textit{Azarenka}'', ``\textit{Nadal}'', and ``\textit{Tomic}'') are mentioned many times in both target posts and their conversations, which reveals why the hashtag is ``\textit{\#AusOpen}''.

To this end, we first put a \textit{post-aware} attention on the conversation encoder with coefficients:
 
\begin{equation} 
\alpha^c_{ij}=\frac{\exp(f_{score}(\mathbf{h}^p_i,\mathbf{h}^c_j))}{\sum_{j'=1}^{|\mathbf{x}^c|} \exp(f_{score}(\mathbf{h}^p_i,\mathbf{h}^c_{j'}))},
\end{equation}
where the alignment score function $f_{score}(\mathbf{h}^p_i,\mathbf{h}^c_j)=\mathbf{h}^p_i\mathbf{W}_{bi-att}\mathbf{h}^c_j$ captures the similarity of the $i$-th word in the target post and the $j$-th word in its conversation. 
Here $\mathbf{W}_{bi-att} \in\mathbb{R}^{d \times d}$ is a weight matrix to be learned.
Then, we compute a context vector $\mathbf{r}^c$ conveying post-aware conversation representations, where the $i$-th value is defined as:   
\begin{equation} 
\mathbf{r}^c_i=\sum_{j=1}^{|\mathbf{x}^c|}\alpha^c_{ij}\mathbf{h}^c_j.
\end{equation}
Analogously, a \textit{conversation-aware} attention on post encoder is used to capture the conversation-aware post representations as $\mathbf{r}^p$. 

\paragraph{Merge Layer.}
Next, to further fuse representations distilled by the bi-attention on each encoder, we design a \textit{merge} layer, a multilayer perceptron (MLP) activated by hyperbolic function:
\begin{equation} \label{vp}
\mathbf{v}^p = \tanh(\mathbf{W}_p[\mathbf{h}^p; \mathbf{r}^c]+\mathbf{b}_p),
\end{equation}
\begin{equation} \label{vc}
\mathbf{v}^c = \tanh(\mathbf{W}_c[\mathbf{h}^c; \mathbf{r}^p]+\mathbf{b}_c),
\end{equation}
where $\mathbf{W}_p,\mathbf{W}_c\in\mathbb{R}^{d\times 2d}$ and $\mathbf{b}_p,\mathbf{b}_c\in\mathbb{R}^{d}$ are trainable parameters.

Note that either $\mathbf{v}^p$ or $\mathbf{v}^c$ conveys the information from both posts and conversations, but with a different emphasis.
Specifically, $\mathbf{v}^p$ mainly retains the contexts of posts with the auxiliary information from conversations, while $\mathbf{v}^c$ does the opposite.
Finally, vectors $\mathbf{v}^p$ and $\mathbf{v}^c$ are concatenated and fed into the decoder for hashtag generation.

\paragraph{Decoder.} 
Given the representations $\mathbf{v}=[\mathbf{v}^p;\mathbf{v}^c]$ produced by our dual encoder with bi-attention, we apply an attention-based GRU decoder to generate a word sequence $\mathbf{y}$ as the hashtag. 
The probability to generate the hashtag conditioned on a target post and its conversation is defined as: 
\begin{equation} 
Pr(\mathbf{y}|\mathbf{x}^p, \mathbf{x}^c)=\prod_{t=1}^{|\mathbf{y}|}Pr(y_t|\mathbf{y}_{<t}, \mathbf{x}^p, \mathbf{x}^c),
\end{equation}
where $\mathbf{y}_{<t}$ refers to $(y_1, y_2,...,y_{t-1})$.

Concretely, when generating the $t$-th word in hashtag, the decoder emits a hidden state vector $\mathbf{s}_t\in\mathbb{R}^d$ and puts a global attention over $\mathbf{v}$. The attention aims to exploit indicative representations from the encoder outputs $\mathbf{v}$ and summarizes them into a context vector $\mathbf{c}_t$ defined as:
\begin{equation} 
\mathbf{c}_t=\sum_{i=1}^{|\mathbf{x}^p|+|\mathbf{x}^c|}\alpha^d_{ti}\mathbf{v}_i,
\end{equation}
\begin{equation} 
\alpha^d_{ti}=\frac{\exp(g_{score}(\mathbf{s}_t,\mathbf{v}_i))}{\sum_{i'=1}^{|\mathbf{x}^p|+|\mathbf{x}^c|} \exp(g_{score}(\mathbf{s}_t,\mathbf{v}_{i'})},
\end{equation}
where $g_{score}(\mathbf{s}_t,\mathbf{v}_i)=\mathbf{s}_t\mathbf{W}_{att}\mathbf{v}_i$ is another alignment function ($\mathbf{W}_{att}\in\mathbb{R}^{d\times d}$) to measure the similarity between $\mathbf{s}_t$ and $\mathbf{v}_i$. 

Finally, we map the current hidden state $\mathbf{s}_t$ of the decoder together with the context vector $\mathbf{c}_t$ to a word distribution over the vocabulary $V$ via: 
\begin{equation} 
Pr(y_t|\mathbf{y}_{<t}, \mathbf{x}^p,  \mathbf{x}^c)=softmax(\mathbf{W}_v[\mathbf{s}_t;\mathbf{c}_t] + \mathbf{b}_v),
\end{equation}
which reflects how likely a word to be the $t$-th word in the generated hashtag sequence.
Here $\mathbf{W}_v\in\mathbb{R}^{V\times2d}$ and $\mathbf{b}_v\in\mathbb{R}^V$ are trainable weights. 

\paragraph{Learning and Inferring Hashtags.}
During the training stage, we apply stochastic gradient descent to minimize the loss function of our entire framework, which is defined as:
\begin{equation} 
\mathcal{L}(\Theta)=-\sum_{n=1}^N\log(Pr(\mathbf{y}_n|\mathbf{x}^p_n,\mathbf{x}^c_n; \Theta)).
\end{equation}
Here $N$ is the number of training instances and $\Theta$ denotes the set of all the learnable parameters. 

In hashtag inference, based on the produced word distribution at each time step, word selection is conducted using beam search. 
In doing so, we generate a ranking list of output hashtags, where the top $K$ hashtags serve as our final output.

%% file: figure_input/framework.tex
\begin{figure}
\centering
\includegraphics[width=7.5cm, height=5.5cm]{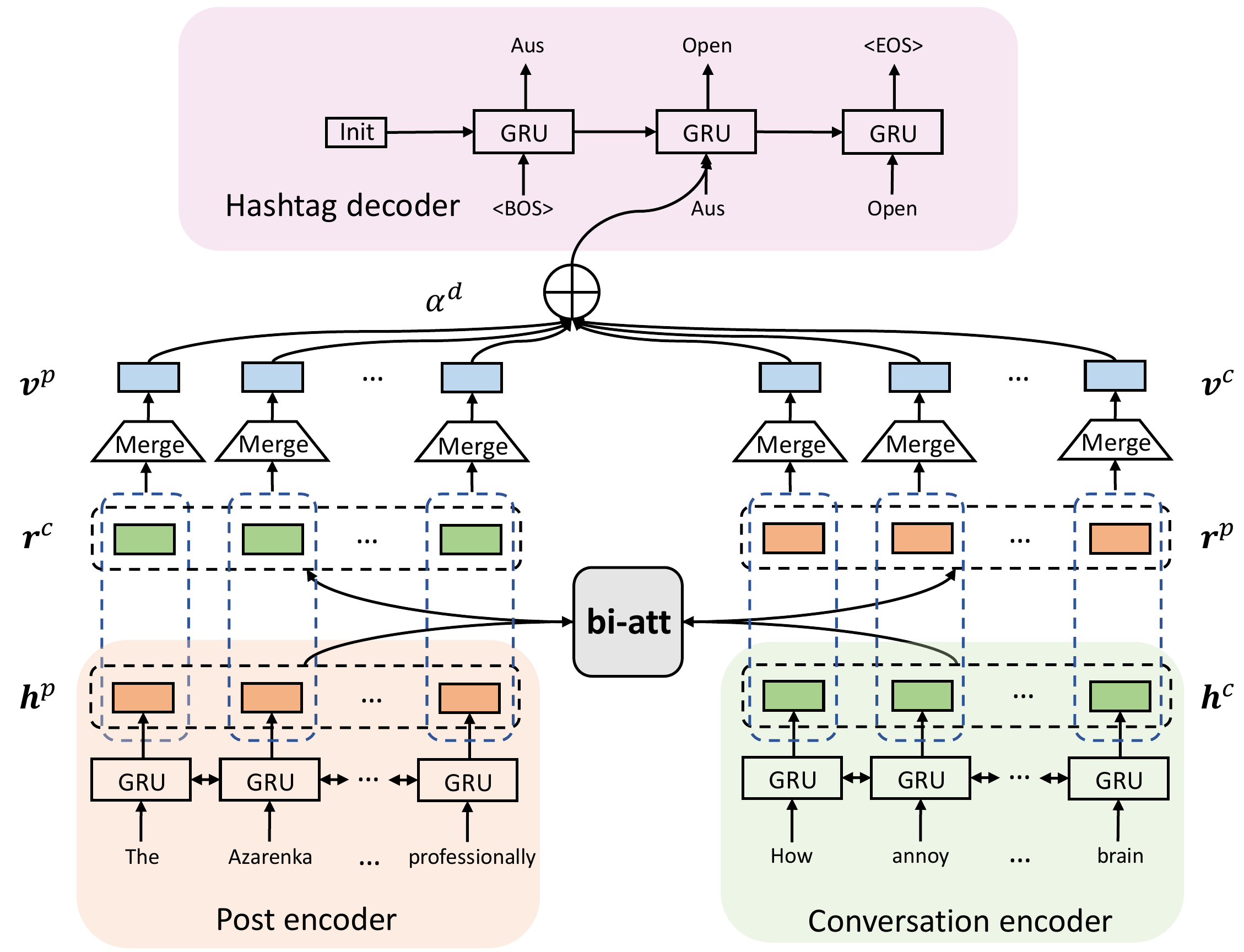}
\caption{Our hashtag generation framework with a dual encoder, including a post encoder and a conversation encoder, where a bi-attention (bi-att) distills their salient features, followed by a merge layer to fuse them. An attentive decoder generates the hashtag sequence.}\label{fig:bi_attn}
\vskip -0.5em
\end{figure}

%% file: sections/exp-setup.tex
\section{Experiment Setup}\label{sec:exp-setup}

\input{tabs/dataset}

Here we describe how we set up our experiments.

\paragraph{Datasets and Statistic Analysis.} 
Two large-scale experiment datasets are \emph{newly collected} from popular microblog platforms: an English Twitter dataset and a Chinese Weibo dataset.
The Twitter dataset was built based on the TREC 2011 microblog track.\footnote{\url{https://trec.nist.gov/data/tweets/}} 
To recover the conversations, we used Tweet Search API
to fetch ``in-reply-to'' relations in a recursive way.
The Weibo dataset was collected from January to August 2014 using Weibo Search API
via searching messages with the trending queries\footnote{ \url{http://open.weibo.com/wiki/Trends/}} as keywords.
For gold-standard hashtags, we take the user-annotated hashtags, appearing before or after a post, as the reference.\footnote{Hashtags in the middle of a post are not considered here as they generally act as semantic elements~\cite{DBLP:conf/emnlp/ZhangWGH16,DBLP:conf/naacl/ZhangLSZ18}.} 
The statistics of our datasets are shown in Table~\ref{tabs:dataset}. 
We randomly split both datasets into three subsets, where $80$\%, $10$\%, and $10$\% of the data corresponds to training, development, and test sets, respectively.

\input{tabs/dataset_hashtags.tex}
\input{figure_input/data_dist.tex}

To further investigate how challenging our problem is, we show some statistics of the hashtags in Table \ref{tabs:dataset_hashtags} and the distributions of hashtag frequency in Figure~\ref{fig:data_dist}.
In Table \ref{tabs:dataset_hashtags}, we observe the large size of hashtags in both datasets. Moreover, Figure \ref{fig:data_dist} indicates that most hashtags only appear a few times.
Given such a large and imbalanced hashtag space, hashtag selection from a candidate list, as many existing methods do, might not perform well.
Table \ref{tabs:dataset_hashtags} also shows that only a small proportion of hashtags appearing in their posts, conversations, and either of them, making it inappropriate to directly extract words from the two sources to form hashtags.

\paragraph{Preprocessing.}
For tokenization and word segmentation, we employed the tweet preprocessing toolkit released by~\citeauthor{DBLP:conf/semeval/BaziotisPD17a}~\shortcite{DBLP:conf/semeval/BaziotisPD17a} for Twitter, and the Jieba toolkit\footnote{\url{https://pypi.python.org/pypi/jieba/}} for Weibo.
Then, for both Twitter and Weibo, we further take the following preprocessing steps: 
First, single-character hashtags were filtered out for not being meaningful. 
Second, generic tags, i.e., links, mentions (@username), and numbers, were replaced with ``URL'' ``MENTION'', and ``DIGIT'',  respectively. 
Third, inappropriate replies (e.g., retweet-only messages) were removed, and the remainder were chronologically ordered to form a sequence as conversation contexts.
Last, a vocabulary was maintained with the $30K$ and $50K$ most frequent words, for Twitter and Weibo, respectively.

\input{tabs/main_exp}

\paragraph{Comparisons.}
For experiment comparisons, we first consider a weak baseline \textsc{Random} that randomly ranks hashtags seen from training data. 
Two unsupervised baselines are also considered, where words are ranked by latent topics induced with the latent Dirichlet allocation topic model (henceforth \textsc{LDA}),
and by their TF-IDF scores (henceforth \textsc{Tf-Idf}). Here for TF-IDF scores, we consider the $N$-gram \textsc{Tf-Idf} ($N\leq 5$).
Besides, we compare with \textit{supervised} models below:

$\bullet$ \textsc{Extractor}: Following \citet{DBLP:conf/naacl/ZhangLSZ18}, we extract phrases from target posts as hashtags via sequence tagging and encode conversations with memory networks~\cite{DBLP:conf/nips/SukhbaatarSWF15}. 

$\bullet$ \textsc{Classifier}: We compare with the state-of-the-art model based on classification~\cite{DBLP:conf/ijcai/GongZ16}, where hashtags are selected from candidates seen in training data. 
Here two versions of their classifier are considered, one only taking a target post as input (henceforth \textsc{Classifier} (\textit{post only})) and the other taking the concatenation of a target post and its conversation as input (henceforth \textsc{Classifier} (\textit{post+conv})).

$\bullet$ \textsc{Generator}: 
A seq2seq generator (henceforth \textsc{Seq2Seq})~\cite{DBLP:conf/nips/SutskeverVL14} is applied to generate hashtags given a target post. 
We also consider its variant augmented with copy mechanism~\cite{DBLP:conf/acl/GuLLL16} (henceforth \textsc{Seq2Seq-copy}), which has proven effective in keyphrase generation~\cite{DBLP:conf/acl/MengZHHBC17} and also takes the post as input. 
The proposed seq2seq with the bi-attention to encode both the post and its conversation is denoted as \textsc{Our model} for simplicity.

\paragraph{Model Settings.} 
We conduct model tunings on the development set based on grid search, where the hyper-parameters that give the lowest objective loss are selected.
For the sequence generation models, the implementations are based on the OpenNMT framework~\cite{DBLP:conf/acl/KleinKDSR17}.
The word embeddings, with dimension set to $200$, are randomly initialized.
For encoders, we employ two layers of Bi-GRU cells, and for decoders, one layer of GRU cell is used. 
The hidden size of all GRUs is set to $300$. 
In learning, we use the Adam optimizer \cite{DBLP:conf/iclr/KingmaB14} with the learning rate initialized to $0.001$.
We adopt the early-stop strategy: the learning rate decreases by a decay rate of $0.5$ till either it is below $1e^{-6}$ or the validation loss stops decreasing.
The norm of gradients is rescaled to $1$ if the $L2$-norm~$>1$ is observed. 
The dropout rate is $0.1$ and the batch size is $64$. 
In inference, we set the beam-size to $20$ and the maximum sequence length of a hashtag to $10$. 

For \textsc{Classifier} and \textsc{Extractor}, lacking publicly available codes, we reimplement the models using Keras.\footnote{\url{https://keras.io/}} 
Their results are reproduced in their original experiment settings. 
For \textsc{LDA}, we employ an open source toolkit lda.\footnote{\url{https://pypi.org/project/lda/}}

\paragraph{Evaluation Metrics.}
Popular \textit{information retrival} evaluation metrics F1 scores at K (F1@K) and mean average precision (MAP) scores~\cite{DBLP:books/daglib/0021593} are reported.
Here, different $K$ values are tested on F1@K and result in a similar trend, so only F1@1 and F1@5 are reported.  
MAP scores are also computed given the top $5$ outputs.    
Besides, as we consider a hashtag as a sequence of words, ROUGE metrics for \textit{summarization} evaluation~\cite{lin:2004:ACLsummarization} are also adopted.
Here, we use ROUGE F1 for the top-ranked hashtag prediction computed by an open source toolkit pythonrouge,\footnote{\url{https://github.com/tagucci/pythonrouge}} with Porter stemmer used for English tweets.
For Weibo posts, scores calculated at the Chinese character level following~\citet{DBLP:journals/cl/LiSWW2018}.
We report the average scores for multiple gold-standard hashtags on ROUGE evaluation.

%% file: tabs/dataset.tex
\begin{table}
\begin{center}

\resizebox{0.49\textwidth}{!}{
\begin{tabular}{|l|c|c|c|c|c|}
\hline
\multirow{2}{*}{\textbf{Datasets}}&   \# of & Avg len &	Avg len & Avg len&\# of tags\\
& posts & of posts & of convs & of tags& per post \\
\hline

Twitter & \hfill{44,793} & \hfill{13.27}  & \hfill{29.94}  & \hfill{1.69}  &   \hfill{1.14} \\

Weibo & \hfill{40,171} & \hfill{32.64}  & \hfill{70.61}  & \hfill{2.70}   & \hfill{1.11} \\

\hline
\end{tabular}
}

\end{center}
\vskip -0.5em
\caption{Statistics of our datasets. Avg len of posts, convs, tags refer to the average number of words in posts, conversations, and hashtags, respectively.} \label{tabs:dataset}

\end{table}

%% file: tabs/dataset_hashtags.tex
\begin{table}
\begin{center}
\resizebox{0.45\textwidth}{!}{
\begin{tabular}{|l|c|ccc|}
\hline
\textbf{Datasets}&$|$Tagset$|$&$\mathcal{P}$&$\mathcal{C}$&$\mathcal{P}\cup\mathcal{C}$\\
\hline
Twitter
&4,188&\hfill{2.72\%}&\hfill{5.58\%} &\hfill{7.69\%}\\

Weibo
&5,027&\hfill{8.29\%}&\hfill{6.21\%}&\hfill{12.52\%}\\

\hline
\end{tabular}
}
\end{center}
\caption{Statistics of the hashtags. $|$Tagset$|$: the number of distinct hashtags. $\mathcal{P}$, $\mathcal{C}$, and $\mathcal{P}\cup\mathcal{C}$: the percentage of hashtags appearing in their corresponding posts, conversations, and the union set of them, respectively.} \label{tabs:dataset_hashtags}
\vskip -0.5em
\end{table}

%% file: figure_input/data_dist.tex
\begin{figure}[ht]
\centering
\includegraphics[scale=0.38, trim=0 0 0 10]{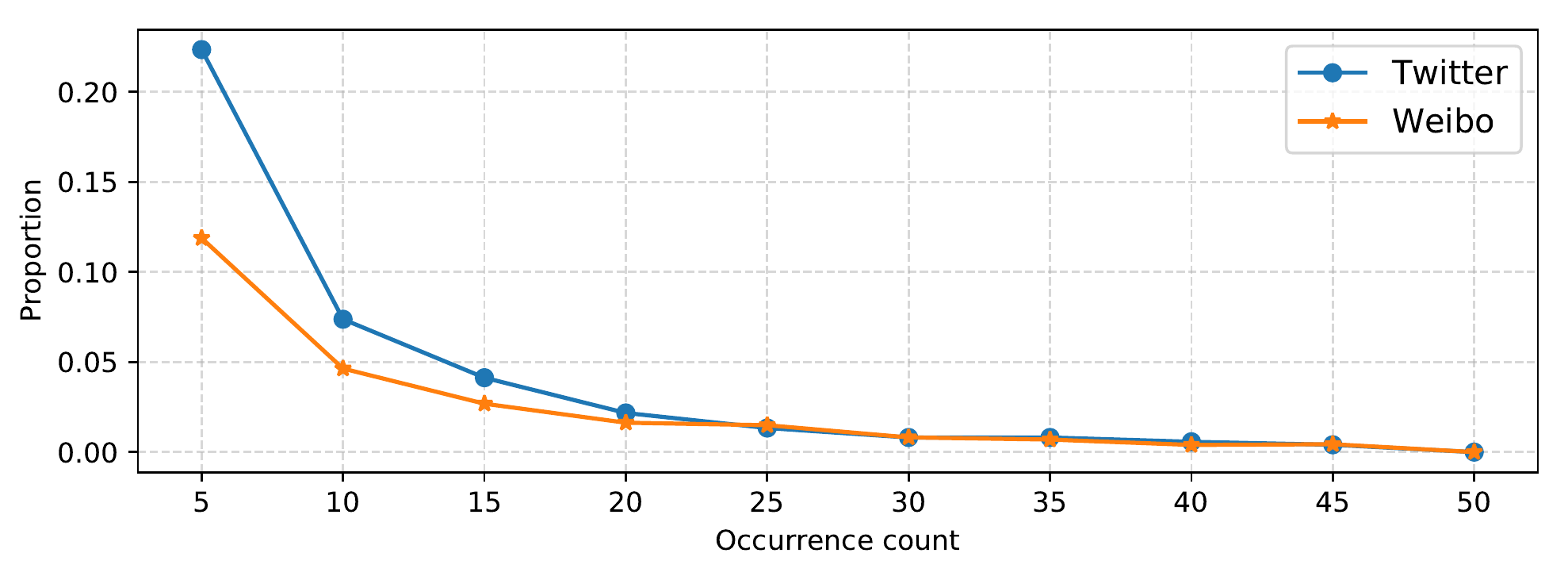}
\vskip -0.5em
\caption{Distribution of hashtag frequency. The horizontal axis refers to the occurrence count of hashtags (shown with maximum $50$ and bin $5$) and the vertical axis denotes the data proportion. }\label{fig:data_dist}
\vskip -0.5em
\end{figure}

%% file: tabs/main_exp.tex
\begin{table*}[ht]
\begin{center}

\resizebox{1.0\textwidth}{!}{  
\begin{tabular}{|l|ccccc||ccccc|}
\hline
\multirow{2}{*}{\textbf{Model}}&\multicolumn{5}{c||}{\textbf{Twitter}}&\multicolumn{5}{c|}{\textbf{Weibo}}\\
\cline{2-11}
&F1@1&F1@5&MAP&RG-1&RG-4 
&F1@1&F1@5&MAP&RG-1&RG-4\\
\cline{2-11}
\hline
\underline{\textbf{Baselines}}&&&&&&&&&&\\
\textsc{Random} 
&0.37&0.63&0.89&0.56&0.16 
&0.43&0.67&0.97&2.14&1.13\\

\textsc{LDA} 
&0.13&0.25&0.35&0.60&-~~~
&0.10&0.86&0.94&3.89&-~~~\\

\textsc{Tf-Idf}
&0.02	&0.02	&0.03	&0.54	&0.14
&0.85	&0.73	&1.30	&8.04	&4.29 \\

\textsc{Extractor} 
&0.44	&-~~~	&-~~~	&1.14	&0.14
&2.53	&-~~~	&-~~~	&7.64	&5.20 \\

\hline
\hline

\underline{\textbf{State of the arts}}&&&&&&&&&&\\

\textsc{Classifier} (\textit{post only}) 
&9.44	&6.36	&12.71	&10.75	&4.00
&16.92	&10.48	&22.29	&25.34	&21.95\\

\textsc{Classifier} (\textit{post+conv}) 
&8.54	&6.28	&12.10	&10.00	&2.47	
&17.25	&11.03	&23.11	&25.16	&22.09\\

\hline
\hline
\underline{\textsc{\textbf{Generators}}}&&&&&&&&&&\\
\textsc{Seq2Seq}
&10.44&6.73&14.00&10.52&4.08
&26.00&14.43&32.74&37.37&32.67\\

\textsc{Seq2Seq-copy}
&10.63	&6.87	&14.21	&12.05	&4.36
&25.29	&14.10	&31.63	&37.58	&32.69\\

\textsc{Our model}  
&\textbf{12.29}* & \textbf{8.29}*&	\textbf{15.94}*	&\textbf{13.73}*	&	\textbf{4.45}

&\textbf{31.96}*&	\textbf{17.39}*&	\textbf{38.79}*&	\textbf{45.03}*&	\textbf{39.73}*\\

\hline
\end{tabular}
}
\end{center}
\caption{Comparison results on Twitter and Weibo datasets (in \%). RG-1 and RG-4 refer to ROUGE-1 and ROUGE-SU4 respectively. The best results in each column are in bold. The ``*'' after numbers indicates significantly better results than all the other models ($p<0.05$, paired t-test). Higher values indicate better performance.
}\label{tabs:main_exp}
\vskip -1em
\end{table*}

%% file: sections/exp-result.tex
\section{Experimental Results}

In this section, we first report the main comparison results in Section~\ref{ssec:main-results}, followed by an in-depth comparative study between classification and sequence generation models
in Section~\ref{ssec:classification-comp}. 
Further discussions are then presented to analyze our superiority and errors in Section~\ref{ssec:discussions}.

\subsection{Main Comparison Results}\label{ssec:main-results}

Table~\ref{tabs:main_exp} reports the main comparison results.
For \textsc{Classifier}, their outputs are ranked according to the logits after a $softmax$ layer.
For \textsc{Extractor}, it is unable to produce ranked hashtags and thus no results are reported for F1@5 and MAP.
For \textsc{LDA}, as it cannot generate bigram hashtags, no results are presented for ROUGE-SU4.
In general, we have the following observations:

$\bullet$ \textbf{\textit{Hashtag annotation is more challenging for Twitter than Weibo.}} 
Generally, all models perform worse on Twitter measured by different metrics.
The intrinsic reason is the essential language difference between English and Chinese microblogs. 
English allows higher freedom in writing, resulting in more variety in Twitter hashtags (e.g., abbreviations are prominent like ``\textit{aus}'' in ``\textit{\#AusOpen}'').
For statistical reasons, Twitter hashtags are more likely to be absent in either posts or conversations (Table~\ref{tabs:dataset_hashtags}), and have a more severe imbalanced distribution (Figure~\ref{fig:data_dist}).

$\bullet$ \textbf{\textit{Topic models and extractive models are ineffective for hashtag annotation}}. 
The poor performance of all baseline models indicates that hashtag annotation is a challenging problem. 
\textsc{LDA} sometimes performs even worse than \textsc{Random} due to its inability to produce phrase-level hashtags.
For extractive models, both \textsc{Tf-Idf} and \textsc{Extractor} fail to achieve good results. 
It is because most hashtags are absent in target posts, as
we see in Table~\ref{tabs:dataset_hashtags} that only $2.72$\% hashtags on Twitter and $8.29$\% on Weibo appear in target posts.
This confirms that extractive models, relying on word selection from target posts, cannot well fit the hashtag annotation scenario. 
For the same reason, copy mechanism fails to bring noticeable improvements for the seq2seq generator on both datasets.

$\bullet$ \textbf{\textit{Sequence generation models outperform other counterparts.}} 
When comparing \textsc{Generators} with other models, we find the former  uniformly achieve better results, showing the superiority to produce hashtags with sequence generation framework. 
Classification models, though as the state of the art, expose their inferiority as they select labels from the large and imbalanced hashtag space (reflected in Table~\ref{tabs:dataset_hashtags} and Figure~\ref{fig:data_dist}). 

$\bullet$ \textbf{\textit{Conversations are useful for hashtag generation.}} 
Among the sequence generation models, \textsc{Our model} achieves the best performance across all the metrics. The observation indicates the usefulness of bi-attention in exploiting the joint effects of target posts and their conversations, which further helps in identifying indicative features from both sources for hashtag generation. 
However, interestingly, incorporating conversations fails to boost the classification performance.
The reason why \textsc{Our model} better exploits conversations than \textsc{Classifier} (\textit{post+conv}) might be that we can attend the indicative features when decoding each word in the hashtag, which is however not possible for classification models (considering hashtags to be inseparable).

\subsection{ Classification vs.  Generation}\label{ssec:classification-comp}

From Table \ref{tabs:main_exp}, we observe that the classifiers outperform topic models and extractive models by a large margin but exhibit generally worse results than sequence generation models. 
Here, we present a thorough study to compare hashtag classification and generation. Four models are selected for comparison: two classifiers, \textsc{Classifier} (\textit{post only}) and \textsc{Classifier} (\textit{post+conv}), and two sequence generation models,  \textsc{Seq2Seq} and \textsc{Our model}.
Below, we explore how they perform to predict rare and new hashtags.

\paragraph{Rare Hashtags.}
According to the hashtag distributions in Figure~\ref{fig:data_dist}, we can see a large proportion of hashtags appearing only a few times in the data. 
To study how models perform to predict such hashtags, in Figure~\ref{fig:freq}, we display their F1@1 scores in inferring hashtags with varying frequency. 
The lower F1 score on less frequent hashtags indicates the difficulty to yield rare hashtags. The reason probably comes from the overfitting issue caused by limited data to learn from. 

We also observe that sequence generation models achieve consistently better F1@1 scores on hashtags with varying sparsity degree, while classification models suffer from the label sparsity issue and obtain worse results.
The better performance of the former might result from the word-by-word generation manner in hashtag generation, which enables the internal structure of hashtags (how words form a hashtag) to be exploited. 

\input{figure_input/f1_freq.tex}

\paragraph{New Hashtags.}
To further explore the extreme situation where hashtags are absent in the training set, we experiment to see how models perform in handling new hashtags. 
To this end, we additionally collect instances tagged with hashtags absent in training data and construct an external test set, with the same size as our original test set.
Considering that classifiers will never predict unseen labels, to ensure comparable performance, we only adopt summarization metrics here for evaluation and report ROUGE-1 F1 scores in Table~\ref{tabs:new_tag}. 

As can be seen, creating unseen hashtags is a challenging task, where unsurprisingly, all models perform poorly on this task.
Nevertheless, sequence generation models perform much better on both datasets, e.g., at least 6.5x improvements over classification models observed on Weibo dataset. 
For Twitter dataset, the improvements are not that large, which confirms again that hashtag annotation on Twitter is more difficult due to the noisier data characteristics.
In particular, compared to \textsc{seq2seq}, \textsc{our model} achieves an additional performance gain in producing new hashtags by leveraging conversations with the bi-attention.

\input{tabs/new_tag_generation}

\subsection{Further Discussions on Our Model}\label{ssec:discussions}

To further analyze our model, we conduct a quantitative ablation study, a qualitative case study, and an error analysis. We then discuss them in turn. 

\paragraph{Ablation Study.}
We report the ablation study results in Table~\ref{tabs:ablation} to examine the relative contributions of the target posts and the conversation contexts. 
To this end, our model is compared with its five variants below: \textsc{Seq2Seq} (\textit{post only}), \textsc{Seq2Seq} (\textit{conv only}), and \textsc{Seq2Seq} (\textit{post+conv}), using standard seq2seq to generate hashtags from their target posts, conversation contexts, and their concatenation, respectively; 
\textsc{Our model} (\textit{post-att only}) and \textsc{Our model} (\textit{conv-att only}), whose decoder only takes $\mathbf{v}^p$ and $\mathbf{v}^c$ defined in Eq.~(\ref{vp}) and Eq.~(\ref{vc}), respectively. 
The results show that solely encoding target posts is more effective than modeling the conversations alone, but exploring their joint effects can further boost the performance, especially combined with a bi-attention mechanism over them.

\input{tabs/ablation}

\paragraph{Case Study.} 
We further present a case study on the target post shown in Table~\ref{tab:intro-example}, where the top five outputs of some comparison models are displayed in Table~\ref{case_study}. 
As can be seen, only our model successfully generates ``\textit{aus open}'', the gold standard. 
Particularly, it not only ranks the correct answer as the top prediction, but also outputs other semantically similar hashtags, e.g., sport-related terms like ``\textit{bbc football}'', ``\textit{arsenal}'', and ``\textit{murray}''. 
On the contrary, \textsc{Classifier} and \textsc{Seq2Seq} tend to yield frequent hashtags, such as ``\textit{just saying}'' and ``\textit{jan 25}''.
Baseline models also perform poorly: \textsc{LDA} produces some common single word, and \textsc{TF-IDF} extracts phrases in the target post, where the gold-standard hashtag is however absent.

\input{tabs/case_study}

To analyze why our model obtains superior results in this case, we display the heatmap in Figure~\ref{fig:bi_attn_vis} to visualize our bi-attention weight matrix $\mathbf{W}_{bi-att}$.
As we can see, bi-attention can identify  the indicative word ``\textit{Azarenka}'' in the target post, via highlighting its other pertinent words in conversations, e.g., ``\textit{Nadal}'' and ``\textit{tennis}''. 
In doing so, salient words in both the post and its conversations can be unveiled, facilitating the correct hashtag ``\textit{aus open}'' to be generated.

\input{figure_input/bi_attn}

\paragraph{Error Analysis.} 
Taking a closer look at our outputs, we find that one type of major errors comes from the unmatched outputs with gold standards, even as a close guess. 
For example, our model predicts ``\textit{super bowl}'' for a post tagged with ``\textit{\#steelers}'', a team in super bowl.
In future work, the semantic similarity should be considered in hashtag evaluation. 
Another primary type of error is caused by the non-topic hashtags, such as ``\textit{\#fb}'' (indicating the messages forwarded from Facebook). 
Such non-topic hashtags cannot reflect any content information from target posts and should be distinguished from topic hashtags in the future.

%% file: figure_input/f1_freq.tex
\begin{figure}
\centering
\includegraphics[width=8.0cm, trim = 0 0 0 10]{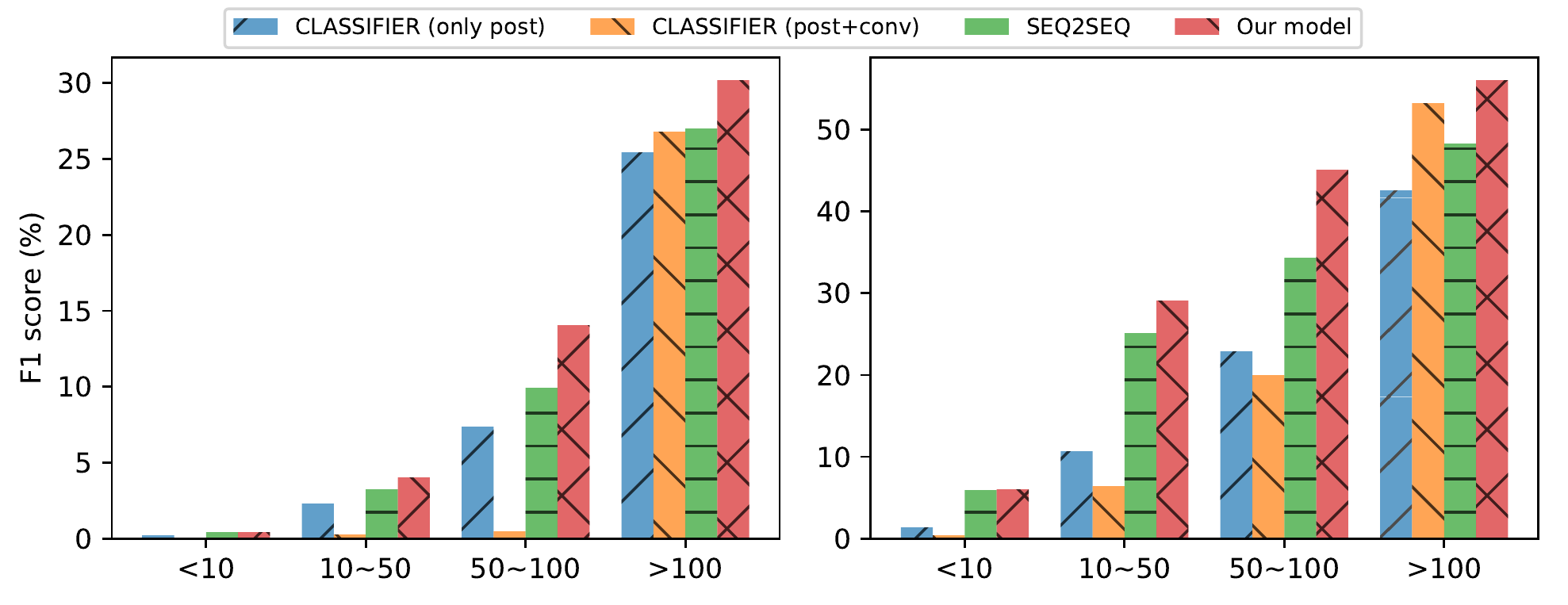}
\caption{F1@1 on Twitter (the left subfigure) and Weibo (the right subfigure) in inferring hashtags with varying frequency.
In each subfigure, from left to right shows the results of \textsc{Classifier} (\textit{post only}), \textsc{Classifier} (\textit{post+conv}), \textsc{Seq2Seq}, and \textsc{Our model}.
Generation models consistently perform better.
}\label{fig:freq}
\end{figure}

%% file: tabs/new_tag_generation.tex
\begin{table}[ht]
\begin{center}
\resizebox{0.45\textwidth}{!}{
\begin{tabular}{|l|r|r|}
\hline
\textbf{Model} &\textbf{Twitter} &\textbf{Weibo}\\
\hline

\textsc{Classifier} (\textit{post only})
&1.15	&1.65\\

\textsc{Classifier} (\textit{post+conv}) 
&1.13	&1.52\\

\textsc{Seq2Seq}  
&1.33	&10.84\\

\textsc{Our model}
&\textbf{1.48}	&\textbf{12.55}\\

\hline
\end{tabular}
}
\end{center}
\caption{ROUGE-1 F1 scores (\%) in producing unseen hashtags. Best results are in bold.}\label{tabs:new_tag}
\end{table}

%% file: tabs/ablation.tex
\begin{table}[H]
\begin{center}
\resizebox{0.45\textwidth}{!}{
\begin{tabular}{|l|r|r|}
\hline
\textbf{Model} &\textbf{Twitter} & \textbf{Weibo}\\
\hline

\textsc{Seq2seq} (\textit{post only})    
&10.44	&26.00\\

\textsc{Seq2seq} (\textit{conv only})   
&6.27 &18.57\\

\textsc{Seq2seq} (\textit{post + conv})    
&11.24	&29.85\\

\textsc{Our model} (\textit{post-att only})
&11.18	&28.67\\

\textsc{Our model} (\textit{conv-att only})
&10.61	&28.06	\\

\textsc{Our model} (\textit{full})
&\textbf{12.29} &\textbf{31.96}\\

\hline
\end{tabular}
}
\end{center}
\vskip -0.5em
\caption{F1@1 scores (\%) for our variants. }\label{tabs:ablation}
\vskip -0.6em
\end{table}


%% file: tabs/case_study.tex
\begin{table}[H]
\centering
\resizebox{0.48\textwidth}{!}{  
\begin{tabular}{|p{2cm}|p{6cm}|}

\hline
\textbf{Model} & \multicolumn{1}{c|}{\textbf{Top five outputs}} \\
\hline
\textsc{LDA}& found; stated; excited; card; apparently \\
\hline
\textsc{TF-IDF}& inappropes; umpire; woman need; azarenka woman; the umpire\\
\hline
\textsc{Classifier} &fail; facebook; just saying; quote; pro choice\\
\hline
\textsc{Seq2seq} & fail; jan 25; yr; eastenders; facebook \\
\hline
\textsc{Our model}& \textit{\textbf{aus open}} ; bbc football ; bbc aus ; arsenal ; murray \\
\hline
\end{tabular}
}
\vskip -0.2em
\caption{Model outputs for the target post in Table~\ref{tab:intro-example}. ``\textit{aus open}'' matches the gold-standard hashtag.}\label{case_study}
\vskip -0.4em
\end{table}

%% file: figure_input/bi_attn.tex
\begin{figure}[H]
\centering
\includegraphics[scale=0.45, width=8cm, height=5cm, trim=0 0 0 0]{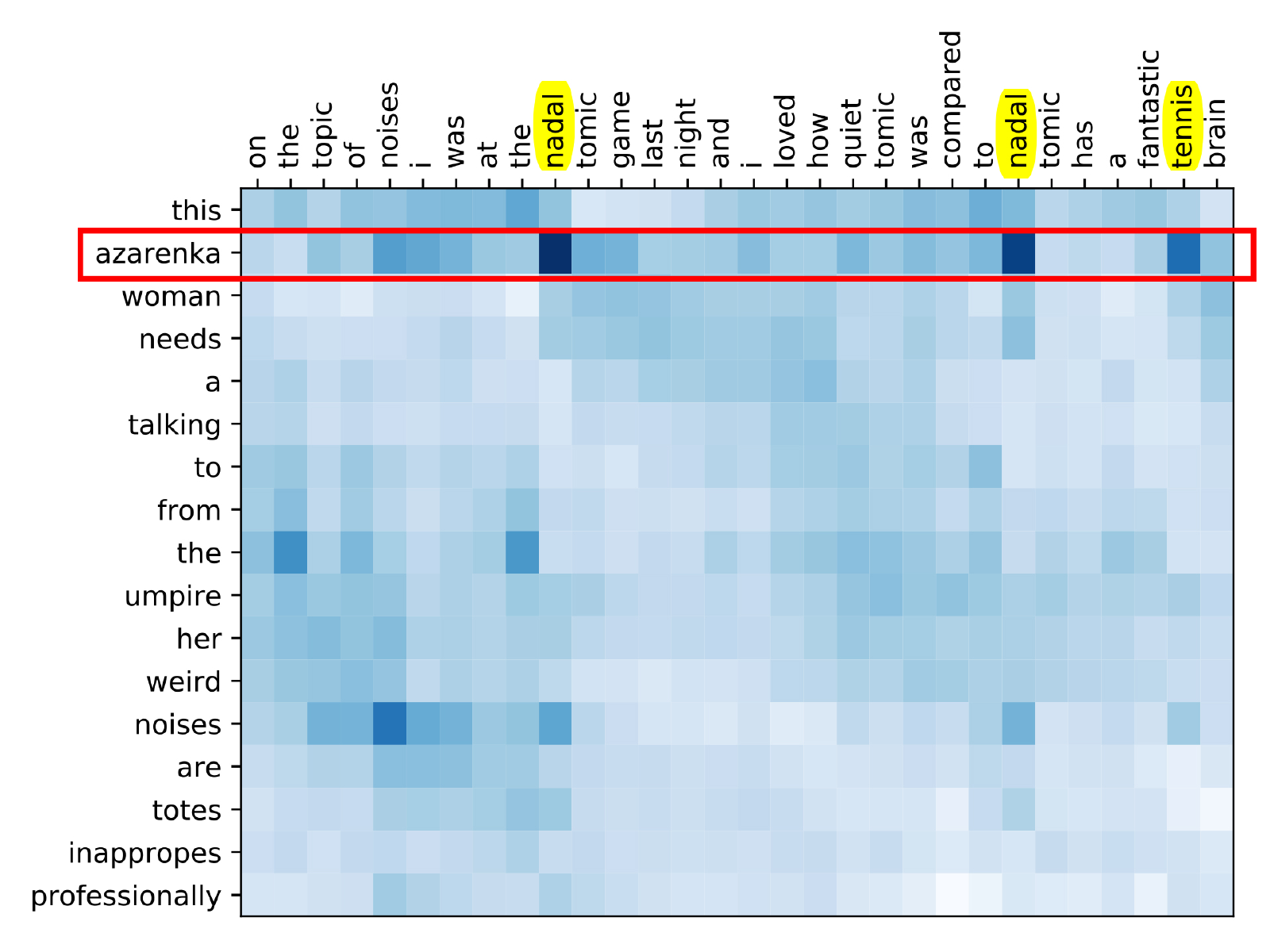}
\caption{Visualization of bi-attention given the input case in Table~\ref{tab:intro-example}. The horizontal axis denotes a snippet of a truncated conversation. The vertical axis shows the target post. Salient words are highlighted.}\label{fig:bi_attn_vis}
\end{figure}

%% file: sections/related-work.tex
\section{Related Work}
Our work mainly builds on two streams of previous
work --- microblog hashtag annotation and neural language generation.

We are in the line of microblog hashtag annotation. 
Some prior work extracts phrases from target posts with sequence tagging models \cite{DBLP:conf/emnlp/ZhangWGH16,DBLP:conf/naacl/ZhangLSZ18}. 
Another popular approach is to apply classifiers and select hashtags from a candidate list
~\cite{DBLP:conf/sigir/HeymannRG08,DBLP:conf/emnlp/WestonCA14,DBLP:conf/sigir/SedhaiS14,DBLP:conf/ijcai/GongZ16,DBLP:conf/coling/HuangZGH16,DBLP:conf/ijcai/ZhangWHHG17}. 
Unlike them, we generate hashtags with a language generation framework, where hashtags in neither the target posts nor the pre-defined candidate list can be created. 
Topic models are also widely applied to induce topic words as hashtags~\cite{DBLP:conf/recsys/KrestelFN09,DBLP:conf/coling/DingZH12,DBLP:conf/www/GodinSNSW13,DBLP:conf/emnlp/GongZH15,DBLP:conf/emnlp/ZhangWGH16}.
However, these models are usually unable to produce phrase-level hashtags, which can be achieved by ours via generating hashtag word sequences with a decoder.

Our work is also closely related to neural language generation, where the encoder-decoder framework~\cite{DBLP:conf/nips/SutskeverVL14} acts as a springboard for many sequence generation models.
In particular, we are inspired by the keyphrase generation studies for scientific articles~\cite{DBLP:conf/acl/MengZHHBC17,DBLP:conf/emnlp/YeW18, DBLP:conf/emnlp/ChenZ0YL18,DBLP:conf/aaai/TgNet_chen}, incorporating word extraction and generation using a seq2seq model with copy mechanism. 
However, our hashtag generation task is inherently different from theirs. 
As we can see from Table~\ref{tabs:main_exp}, it is suboptimal to directly apply keyphrase generation models on our data. 
The reason mostly lies in the informal language style of microblog users in writing both target posts and their hashtags. 
To adapt our model on microblog data, we explore the effects of conversation contexts on hashtag generation, which has never been studied in any prior work before.

%% file: sections/conclusion.tex
\section{Conclusion}
We have presented a novel framework of hashtag generation via jointly modeling of target posts and conversation contexts. To this end, we have proposed a neural seq2seq model with bi-attention over a dual encoder for capturing indicative representations from the two sources. Experimental results on two newly collected datasets have demonstrated that our proposed model significantly outperforms existing state-of-the-art models. Further studies have shown that our model can effectively generate rare and even unseen hashtags.

%% file: sections/ack.tex
\section*{Acknowledgements}
This work is supported by the Research Grants Council of the
Hong Kong Special Administrative Region, China (No. CUHK 14208815 and No. CUHK 14210717 of the General Research Fund). We thank NAACL reviewers for their insightful suggestions on various aspects of this work.